\documentclass[10pt]{article} % For LaTeX2e

\usepackage[preprint]{tmlr}
\usepackage{amsmath,amsfonts,bm}

\def\eqref#1{equation~\ref{#1}}
\def\1{\bm{1}}

\def\ve{{\bm{e}}}

\def\vx{{\bm{x}}}

\DeclareMathAlphabet{\mathsfit}{\encodingdefault}{\sfdefault}{m}{sl}
\SetMathAlphabet{\mathsfit}{bold}{\encodingdefault}{\sfdefault}{bx}{n}

\def\gH{{\mathcal{H}}}

\def\gL{{\mathcal{L}}}

\def\gN{{\mathcal{N}}}

\def\gS{{\mathcal{S}}}
\def\gT{{\mathcal{T}}}

\newcommand{\R}{\mathbb{R}}

\usepackage[utf8]{inputenc} % allow utf-8 input
\usepackage[T1]{fontenc}    % use 8-bit T1 fonts
\definecolor{linkColor}{rgb}{0.18,0.39,0.62}
\usepackage[colorlinks=true,linkcolor=linkColor,citecolor=linkColor,filecolor=linkColor,urlcolor=linkColor]{hyperref}       % hyperlinks
\usepackage{url}            % simple URL typesetting
\usepackage{booktabs}       % professional-quality tables
\usepackage{amsfonts}       % blackboard math symbols
\usepackage{nicefrac}       % compact symbols for 1/2, etc.
\usepackage{microtype}      % microtypography
\usepackage{xspace}
\usepackage{subfig}
\usepackage{enumitem}

\usepackage{multirow}
\usepackage{makecell}
\usepackage{amsmath}
\usepackage{capt-of}
\usepackage{tabularx}
\usepackage{epsfig}
\usepackage{amssymb}
\usepackage{amsfonts}
\usepackage{booktabs}
\usepackage{scalerel}
\usepackage{listings}
\usepackage{varwidth}
\usepackage[export]{adjustbox}
\usepackage{tikz}
\usetikzlibrary{tikzmark}

\usepackage{stmaryrd}
\usepackage{bbm}
\usepackage{wrapfig}
\usepackage{pifont}

\definecolor{deepblue}{rgb}{0,0,0.5}
\definecolor{officeblue}{RGB}{0,102,204}
\definecolor{deepred}{rgb}{0.6,0,0}
\definecolor{deepgreen}{rgb}{0,0.5,0}
\definecolor{mybrickred}{RGB}{182,50,28}

\definecolor{fillcolor}{RGB}{216,217,252}

\usepackage{pifont}% http://ctan.org/pkg/pifont
\newcommand{\cmark}{\ding{51}\xspace}%
\newcommand{\xmark}{\ding{55}\xspace}%

\newcommand\beit{\textsc{BEiT}}
\newcommand\beitii{\textsc{BEiT v2}}

\newcommand\our{\textsc{MaskDistill}}
\newcommand\vqkd{\textsc{VQ-KD}}

\newcommand{\ie}{\textit{i}.\textit{e}.}
\newcommand{\eg}{\textit{e}.\textit{g}.}

\newcommand\deitiii{DeiT \uppercase\expandafter{\romannumeral3}}

\definecolor{teacher}{RGB}{242, 152, 119}
\definecolor{student}{RGB}{3, 115, 140}
\definecolor{human}{RGB}{242, 82, 68}
\title{A Unified View of Masked Image Modeling}

\author{%
{Zhiliang Peng$^{1}$\thanks{~Contribution during internship at Microsoft Research.},~~Li Dong$^{2}$,~~Hangbo Bao$^{2}$,~~Qixiang Ye$^{1}$,~~Furu Wei$^{2}$} \\
\addr University of Chinese Academy of Sciences$^{1}$ \\
Microsoft Research$^{2}$ \\
\url{https://aka.ms/unimim}
}

\def\month{MM}  % Insert correct month for camera-ready version
\def\year{YYYY} % Insert correct year for camera-ready version
\def\openreview{\url{https://openreview.net/forum?id=XXXX}} % Insert correct link to OpenReview for camera-ready version

\begin{document}

\maketitle

\begin{abstract}
Masked image modeling has demonstrated great potential to eliminate the label-hungry problem of training large-scale vision Transformers, achieving impressive performance on various downstream tasks.
% Motivation
% However, the research on this line lacks a systemic guideline.
%
In this work, we propose a unified view of masked image modeling after revisiting existing methods.
Under the unified view, we introduce a simple yet effective method, termed as \our{}, which reconstructs normalized semantic features from teacher models at the masked positions, conditioning on corrupted input images.
Experimental results on image classification and semantic segmentation show that \our{} achieves comparable or superior performance than state-of-the-art methods.
When using the huge vision Transformer and pretraining 300 epochs, \our{} obtains 88.3\% fine-tuning top-1 accuracy on ImageNet-1k (224 size) and 58.8\% semantic segmentation mIoU metric on ADE20k (512 size).
% Code is enclosed in the supplementary materials.
The code and pretrained models will be available at \url{https://aka.ms/unimim}.
\end{abstract}

\section{Introduction}

In recent years, Transformer architectures have shown promising results in the natural language processing field~\citep{transformer} and computer vision field~\citep{vit}.
Transformer, in the process of scaling up, is easy to overfit the small datasets and tends to demand more and more data.
In NLP, self-supervised pretraining methods based on language modeling~\citep{gpt,bert,unilm}, have successfully addressed this problem.
Motivated by masked language modeling, \beit{}~\citep{beit} proposes masked image modeling (MIM) to relieve the label-hungry problem of vision Transformers (ViT; \citealt{vit}), which shows impressive results in learning visual representations. 

MIM is conceptually simple: models accept the corrupted input image and predict the target of the masked content.
Take the pioneering work \beit{}~\citep{beit} as an example, the encoder accepts corrupted image patches as input and then predicts the corresponding discrete visual tokens from the tokenizer~\citep{dalle} at the masked positions. After that, the main difference between previous work lies in the architecture design~\citep{mae,cae} and reconstruction targets~\citep{mae,swinv2,maskfeat,mvp,data2vec}.

\begin{figure}[t]
\begin{center}
\begin{tabular}{c}
\includegraphics[width=1\textwidth]{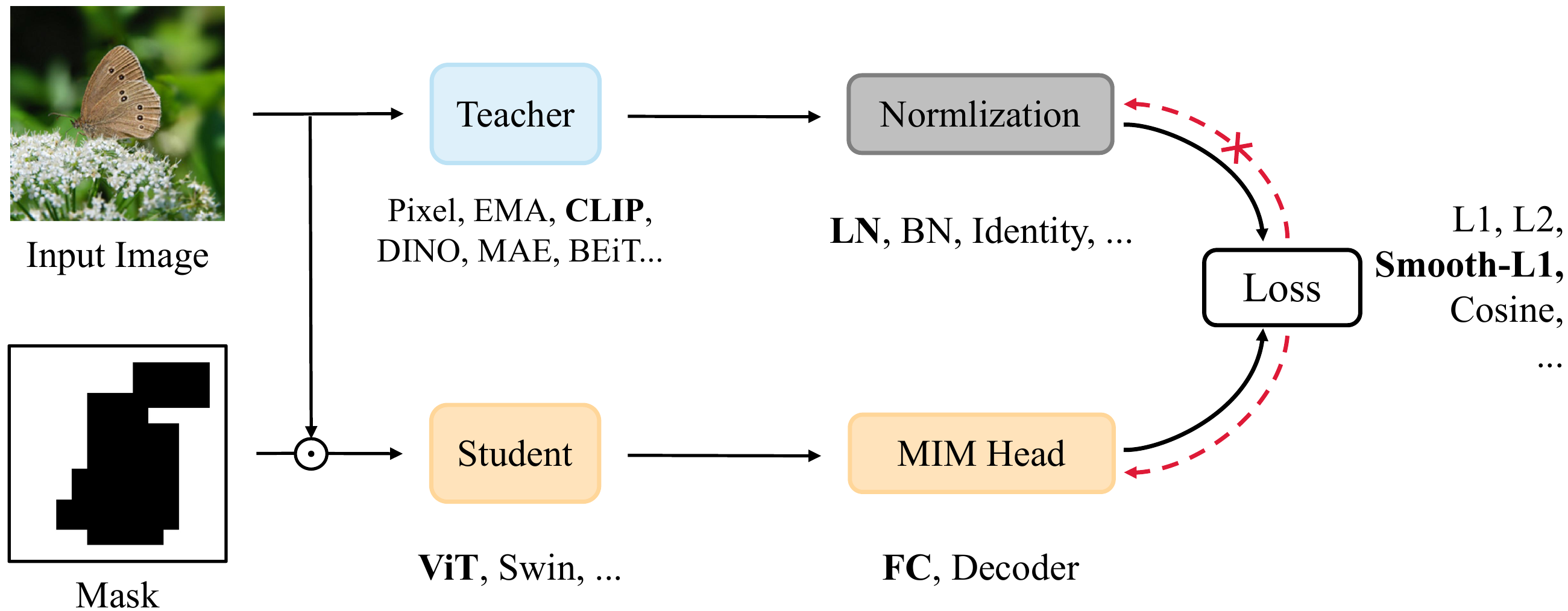}
\end{tabular}
\end{center}
\caption{Unified view of the masked image modeling framework. The \textbf{bold} text denotes the default ingredients of \our{}.
}
\label{fig:maskdistill}
\vspace{-1em}
\end{figure}

In this work, we provide a unified view of masked image modeling, as illustrated in Equation~\ref{eq:unify} and Figure~\ref{fig:maskdistill}: a teacher model, a normalization layer, a student model, a MIM head, and a proper loss function.
According to it, we conduct a systemic comparison of the recent MIM works and present it in Table~\ref{tbl:unify}. 
The most significant difference is the teacher model selection,
$e.g.$, pixel, tokenizers, pretrained models, and the momentum updated teacher.

Under this unified view, we induce a simple yet effective method, named \our{}. 
As shown in Figure~\ref{fig:maskdistill}, the ingredients of \our{} contain a teacher model based on CLIP~\citep{clip}, a fully-connection layer MIM head, layer normalization for target feature, and the Smooth-$\ell_1$ loss function.
Compared to existing methods in Table~\ref{tbl:unify}, \our{} is loyal to the most straightforward design, but shows impressive results.
% Compared to the latest knowledge distillation method~\citep{fd_clip}, \our{} shows superior performance and scalability.
Compared to knowledge distillation, \our{} pays more attention to extrapolating the masked patches rather than mimicking the target features.

We conduct MIM pretraining on ImageNet-1k~\citep{imagenet} for base-, large- and huge-size ViTs. 
After that, we evaluate pretraining models on downstream visual tasks, image classification on ImageNet-1k, and semantic segmentation on ADE20k~\citep{ade20k}. With the large-size CLIP teacher, \our{} using ViT-H/14 can achieve 88.3\% accuracy on ImageNet-1k and 58.8\% mIoU on ADE20k, by pretraining 300 epochs.

% Surprisingly, we find that using the teacher trained with image-image contrastive loss achieves comparable performance to using the teacher supervised by image-text pair contrastive loss on academically accessible datasets.
% That is, only using language-guided visual representation as the target for MIM does not work better than not using language information.
% Therefore, how to squeeze the potential of language to boost visual representation is still an open issue.

The contributions of this study are summarized as follows:
\begin{itemize}[leftmargin=1.5em]
\item We provide a unified view of masked image modeling:  a teacher model, a normalization layer, a student model, a MIM head, and a proper loss function.
\item We propose a simple yet effective method, termed as \our{}.
\item We conduct extensive experiments on downstream tasks including ImageNet fine-tuning and semantic segmentation. Experimental results show that the proposed approach improves performance across various settings.
\end{itemize}

\section{A Unified View of Masked Image Modeling}

\begin{table}[t]
\centering
\caption{Systemic comparisons of masked image modeling methods from a unified view.
}
\label{tbl:unify}
\resizebox{\linewidth}{!}{ %resize
\begin{tabular}{@{}lccccc@{}}
\toprule
\bf Methods & \bf Teacher $\gT$ & \bf Student $\gS$ & \bf MIM Head $\gH$ & \bf Normalization $\gN$ & \bf Loss Function $\gL$ \\
\midrule
\multicolumn{5}{l}{\textit{Low-level pixel / feature}} \\
ViT~\citep{vit} & Pixel & ViT & FC & / & N/A  \\
MAE~\citep{mae} & Pixel & ViT & Decoder & LayerNorm & $\ell_2$ \\
SimMIM~\citep{swinv2} & Pixel  & Swin & FC & / &  $\ell_1$ \\
MaskFeat~\citep{maskfeat} & HOG  & ViT & FC & / & $\ell_2$ \\ 
Ge$^2$-AE~\citep{Ge2-AE} & Pixel\&Frequency  & ViT & Decoders & / & $\ell_2$ \\
ConvMAE~\citep{ConvMAE} & Pixel & Hybrid ViT & Decoder & LayerNorm & $\ell_2$  \\
HiViT~\citep{HiViT} & Pixel & HiViT & Decoder & LayerNorm & $\ell_2$  \\
GreenMIM~\citep{GreenMIM} & Pixel  & Swin & Decoder & LayerNorm &  $\ell_2$ \\
\midrule
\multicolumn{5}{l}{\textit{High-level feature}} \\
\beit{}~\citep{beit} & dVAE & ViT & FC & / & CrossEntropy \\
CAE~\citep{cae} & dVAE & ViT & Decoder & / & CrossEntropy \\
SplitMask~\citep{splitmask} & dVAE & ViT & Decoder & / & InfoNCE\&CrossEnt. \\
PeCo~\citep{peco} & VQGAN & ViT & FC & / & CrossEntropy \\
BEiT v2~\citep{beitv2} & VQ-KD & ViT & FC & / & CrossEntropy  \\ 
MaskFeat~\citep{maskfeat} & DINO & ViT & FC  & ($\ell_2$) & Cosine \\
MVP~\citep{mvp} & CLIP & ViT & FC  & ($\ell_2$) & Cosine \\
MILAN~\citep{MILAN} & CLIP & ViT & Decoders & $\ell_2$-Norm & $\ell_2$ \\
MimCo~\citep{MimCo} & MoCov3 & ViT & FC  & / & InfoNCE \\
data2vec~\citep{data2vec} & EMA & ViT & FC   & LayerNorm & Smooth-$\ell_1$ \\ 
MSN~\citep{MSN} & EMA & ViT & FC & / & CrossEntropy \\ 
SIM~\citep{sim} & EMA & ViT & Decoder & BatchNorm & UniGrad loss \\ 
SdAE~\citep{SdAE} & EMA & ViT & Decoder & LayerNorm & Cosine \\ 
ConMIM~\citep{ConMIM} & EMA & ViT & FC & BatchNorm & InfoNCE \\ 
ExtreMA~\citep{ExtreMA} & EMA & ViT & CrossAtt & LayerNorm & Cosine \\
BootMAE~\citep{BootMAE} & EMA\&Pixel & ViT & Decoders & LayerNorm & $\ell_2$ \\ 
\midrule
\bf \our{} (Ours) & CLIP & ViT & FC & LayerNorm & Smooth-$\ell_1$ \\
\bottomrule
\end{tabular}
} %resize
% \vspace{1em}
\end{table}

In this section, we provide a unified view of the masked image modeling (MIM) task: a teacher model $\gT$, a normalization layer $\gN$, a student model $\gS$, a MIM head $\gH$, and an objective function $\gL$ that measures the distance between the representation of the teacher model $\gT$ and that of the student model $\gS$.
The pretraining task can be unified as:
\begin{align}
\text{MIM} = \gL(\gN(\gT(I_{\text{full}})), \gH(\gS(I_{\text{masked}})))
\label{eq:unify}
\end{align}
where $I_{\text{full}}$ and $I_{\text{masked}}$ denote the full (original) image and the masked image respectively.
According to Equation~\ref{eq:unify}, we summarize the recent popular MIM works in Table~\ref{tbl:unify}. 

1) \textit{Teacher models} $\gT$. 
According to the semantic information of target, we split them into two groups: \textit{low-level} and \textit{high-level} target.
For the low-level target, ViT~\citep{vit}, MAE~\citep{mae}, SimMIM~\citep{swinv2}, ConvMAE~\citep{ConvMAE}, HiViT~\citep{HiViT} and GreenMIM~\citep{GreenMIM} utilize the original or normalized pixels as the MIM target.
MaskFeat~\citep{maskfeat} introduces the feature descriptor HOG~\citep{hog} as the regression target.
And Ge$^2$-AE regresses pixel and frequency from 2D-Discrete Fourier Transform in parallel.
As for high-level target, \beit{}~\citep{beit}, CAE~\citep{cae}, SplitMask~\citep{splitmask}, PeCo~\citep{peco} and \beitii{}~\citep{beitv2} predict the discrete tokens (instantiated as code in the visual tokenizer~\citep{dalle,vqgan,beitv2}.
MaskFeat~\citep{maskfeat} proposes to directly regress the pretrained model ($e.g.$, DINO~\citep{dino} and DeiT~\citep{deit}).
MVP~\citep{mvp} extends the pretrained model to the multimodal pretrained model CLIP~\citep{clip}. 
Moreover, following the BYOL paradigm~\citep{byol}, data2vec~\citep{data2vec}, MSN~\citep{MSN},  ConMIM~\citep{ConMIM}, SIM~\citep{sim} and BootMAE~\citep{BootMAE} construct the regression target from the momentum updated teacher to boost itself online.

2) \textit{Student models} $\gS$. MIM task is suitable for the models root in attention interaction, like ViT~\citep{vit}, Swin Transformers~\citep{swinv2}, and some variants~\citep{ConvMAE,HiViT}. Because backbone architecture is not the primary focus of this study, we choose the vanilla ViT~\citep{vit} as the analytical anchor.

3) \textit{MIM Heads} $\gH$. \beit{}~\citep{beit} uses a simple fully-connection (FC) layer as the task head to generate prediction at the masked positions. 
MAE~\citep{mae} introduces a decoder to decouple the masked prediction task from the encoder. 
% This special design shows faster speed and superior performance, which is followed by many works (Decoder in Table~\ref{tbl:unify}).
In fact, the aim of the decoder in MAE is still to predict the target pixel at the masked positions.
Therefore, we consider the decoder as a MIM head in Table~\ref{tbl:unify}. 
And this decoupling decoder is adopted by many recent works~\citep{Ge2-AE,ConvMAE,HiViT,cae,splitmask,sim,BootMAE}.

4) \textit{Normalization Layers $\gN$}. 
% BYOL~\citep{byol} uses batch normalization to avoid collapse in presence of the momentum teacher. 
MAE~\citep{mae} also introduces per-patch normalized pixels ($i.e.$, layer normalization without affine transformation) as the target to boost local pixels contrast, resulting in better performance.
Meanwhile, normalization is usually applied for avoiding feature collapse in methods based on contrastive learning~\citep{byol,simclr}.
Similarly, EMA-based MIM methods~\citep{sim,data2vec,ConMIM} adopt various normalization methods to stabilize training as well as boost performance.
There is no collapse issue when the teacher is pixels or frozen models by default.

5) \textit{Loss functions} $\gL$.
When the target is pixel or feature, $\ell_1$ or $\ell_2$ losses are appropriate for feature regression.
When the target is discrete tokens, the cross entropy loss is the primary  choice.
Notably, after applying layer normalization, the variance of target feature rises, resulting in volatile loss, whereas Smooth-$\ell_1$ loss is a trade-off between $\ell_1$ and $\ell_2$, performing more stable.
Of course, cosine similarity loss is also an alternative choice.

% 4) \textit{Loss functions} $\gL$ \textit{\&} \textit{Normalization}.

% some conclusions
From Table~\ref{tbl:unify}, one can find that the main difference is the teacher models: pixel, momentum-updated teachers, and pretrained models.
Pixel is easy to access but struggles with low-level semantic knowledge.
Momentum-updated teachers do not need extra models or datasets but tend to suffer from the collapse issue.
Pretrained models are off-the-shelf and contain more rich semantic information than pixels, but how to prepare a high-quality teacher model is an essential problem.

\section{Masked Distillation}

% rephrase from minilm, introduce KD
Knowledge distillation~\citep{hinton2015distilling} has shown to be a promising approach for compressing a large model (referred to as the teacher model) into a small model (referred to as the student model), which utilizes much fewer parameters and computations while attaining comparable results on downstream tasks.

Based on the unified view, we offer a simple yet effective method, named \our{}, to distill a student model in a masked image modeling fashion.
However, our purpose is not to compress the teacher model $\gT$ into the student model $\gS$, but to boost $\gS$ to outperform $\gT$.
We instantiate the student model $\gS$ as ViT~\citep{vit} for comparison with others.

% reuse the definition in beitv2.
% TODO, rephrasing it
Specially, given the input image $\vx \in \R^{H \times W \times C}$, where $(H, W)$ is the resolution and $C$ is the number of image channels, the student $\gS$ first divides $\vx$ into $N$ non-overlapping patches $\{\vx^{p}_{i} \}_{i=1}^{N}$ and then linear projects it into patch embeddings $\{\ve^{p}_{i} \}_{i=1}^{N}$. 
Following that, we select roughly 40\% of the image patch embeddings to be masked, in a block-wise strategy~\citep{beit}.
Denoting the masked position set as $\mathcal{M}$, we use a shared learnable embedding $\ve_{M}$ to replace the original patch embeddings $\ve^p_{i}$ if $i \in \mathcal{M}$. After that, we get the masked sequence:
\begin{align}
\ve_{i}^{\mathcal{M}} = \delta(i \in \mathcal{M}) \odot \ve_{M} + (1 - \delta(i \in \mathcal{M})) \odot \ve^{p}_{i},
\label{eq:x_mask}
\end{align}
where $\delta(\cdot)$ is the indicator function. 
Subsequently, we prepend a learnable class token $\ve_{\texttt{CLS}}$ and add the learnable positional embeddings, and then feed those into stacked transformer blocks. Lastly, a masked image modeling head (usually instantiate as a fully-connected layer) is applied for predicting feature $\textbf{O} \in \R^{(N+1)\times D}$, where $D$ is the dimension of target features.

Given a pretrained teacher model $\gT$, like DINO~\citep{dino} and CLIP~\citep{clip}, the same image $\vx$ is fed into $\gT$ to get the target feature $\{t_{i}\}^{N}_{i=1}$ patch-to-patch. 
To ensure that the output resolution of $\gS$ and $\gT$ is the same, the input resolution for $\gT$ should be adjusted.
Finally, the training objective of \our{} can be formulated as:
\begin{align}
\gL_{\our{}} = -\sum_{i \in \mathcal{M}}\mathrm{log}(t_{i}(x)|\vx_{i}^{p}) =
\frac{1}{|\mathcal{M}|}\sum_{i \in \mathcal{M}}\text{Smooth-}\ell_1(o_i, LN(t_i)),
\label{eq:mim_objective}
\end{align}
where $LN$ is the layer normalization without affine transformation.

\begin{table}[t]
\centering
\caption{Fine-tuning results on ImageNet-1K and ADE20k.}
\label{tbl:results:cls:imagenet}
% \resizebox{\linewidth}{!}{ %resize
\begin{tabular}{@{}lcccc@{}}
\toprule
\bf \multirow{2}{*}{Methods} & \bf Pretraining & \bf \multirow{2}{*}{Supervision} & \bf Classification & \bf Segmentation \\
& \bf Epochs &  &\bf Top-1 Acc (\%) &\bf mIoU (\%) \\
\midrule
\multicolumn{5}{l}{\textit{Base-size models (ViT-B/16)}} \\
% ViT$_{384}$~\citep{vit} & -  & Label  & 77.9 & - \\
% DeiT~\citep{deit} & -  & Label & 81.8 & - \\
% \deitiii{}~\citep{deit3} & 800  & Label  & 83.8 & 49.3 \\
% MoCo v3~\citep{mocov3}  & 300  & EMA  & 83.2 & 47.3\\
\beit{}~\citep{beit}    & 800  & DALL-E  & 83.2 & 45.6 \\
% DINO~\citep{dino}       & 400 & EMA & 83.6 & 46.8 \\
MAE~\citep{mae}         & 1600  & Pixel & 83.6 & 48.1 \\
CAE~\citep{cae}         & 1600  & DALL-E & 83.9 & 50.2 \\
SdAE~\citep{SdAE} & 300  & EMA & 84.1 & 48.6 \\
SIM~\citep{sim} & 1600 & EMA & 83.8 & N/A \\
% FD-DINO~\citep{fd_clip} & 300  & DINO & 83.8 & 47.7 \\
% MaskFeat~\citep{maskfeat} & 300  & DINO & 84.0 & N/A \\
MaskFeat~\citep{maskfeat} & 1600  & HOG & 84.0 & N/A \\
PeCo~\citep{peco} & 300  & VQGAN & 84.1 & 46.7 \\
PeCo~\citep{peco} & 800  & VQGAN & 84.5 & 48.5 \\
data2vec~\citep{data2vec} & 800  & EMA & 84.2 & N/A \\
CLIP~\citep{clip} & -  & Text  & 84.9 & 51.1 \\
MVP~\citep{mvp} & 300  & CLIP-B & 84.4 & 52.4 \\
% FD-CLIP~\citep{fd_clip} & 300  & CLIP-B & 84.9 & 52.8 \\
\beitii{}~\citep{beitv2} & 1600 & \vqkd{} & 85.5 & 53.1 \\
% \midrule
\our{} (ours) & 300  &   CLIP-B & 85.0 & 53.8 \\
\our{} (ours) & 800  & CLIP-B & 85.5 & 54.3  \\
\midrule
\multicolumn{5}{l}{\textit{Large-size models (ViT-L/16)}} \\
% \deitiii{}~\citep{deit3} & 800  & Label & 84.9 & 51.5 \\
MaskFeat~\citep{maskfeat} & 1600  & HOG & 85.7 & N/A \\
MAE~\citep{mae}         & 1600  & Pixel & 85.9 & 53.6\\
CAE~\citep{cae}         & 1600  & DALL-E & 86.3 & 54.7\\
data2vec~\citep{data2vec} & 1600  & EMA & 86.6 & N/A \\
\beitii{}~\citep{beitv2} & 1600  & \vqkd{} & 87.3 & 56.7 \\
MILAN~\citep{MILAN} & 400 & CLIP-B & 86.7 & 55.3 \\
\our{} (ours) & 300  & CLIP-B & 86.8 & 56.3 \\
\our{} (ours) & 800  & CLIP-B & 87.1 & 56.5  \\
% \midrule
% \multicolumn{7}{l}{\textit{Scaling up to larger teacher, CLIP-L/14}} \\
% \our{} (ours) & ViT-B/16 & 300  & CLIP-L & 50 & 85.3 & 54.3 \\
% \our{} (ours) & ViT-L/16 & 300  & CLIP-L & 50 & 87.6 & 57.9 \\
% \our{} (ours) & ViT-H/14 & 300  & CLIP-L & 30 & 88.3 & 58.8 \\
\bottomrule
\end{tabular}
% } % resize
% ${\dag}$ is our implementation for fine-tuning with the official ViT-B/16 checkpoint.
\end{table}

\begin{table}[t]
\centering
\caption{Fine-tuning results on ImageNet-1K and ADE20k. 
The teacher is CLIP ViT-L/14.
% ${\dag}$ is our implementation for fine-tuning with the official ViT-B/16 checkpoint.
}
\label{tbl:results:cls:imagenet_2}
% \resizebox{\linewidth}{!}{ %resize
\begin{tabular}{@{}lccccc@{}}
\toprule
\bf \multirow{2}{*}{Methods} & \bf Model & \bf Pretraining & \bf \multirow{2}{*}{Supervision} & \bf Classification & \bf Segmentation \\
&\bf Size &\bf Epochs &  &\bf Top-1 Acc (\%) &\bf mIoU (\%) \\
\midrule
\multicolumn{6}{l}{\textit{Scaling up to larger teacher, CLIP ViT-L/14}} \\
\our{} (ours) & ViT-B/16 & 300  & CLIP-L & 85.3 & 54.3 \\
\our{} (ours) & ViT-L/16 & 300  & CLIP-L & 87.6 & 57.9 \\
\our{} (ours) & ViT-H/14 & 300  & CLIP-L & 88.3 & 58.8 \\
\bottomrule
\end{tabular}
% } % resize
\end{table}
 
\section{Experiments}

We perform pretraining and then evaluate finetuning performance on various downstream tasks, such as image classification and semantic segmentation.
Moreover, we conduct ablation studies to compare the contributions of different design choices.

% \subsection{\our{} Implementation and Evaluation}
\subsection{Setup}
For all pretraining experiments, we only use the ImageNet-1k dataset~\citep{imagenet} contains 1.28M images. 
We adopt the block masking strategy to corrupt the input images for the student model, but keep the full images for the teacher, to construct the asymmetric informational bottleneck.
All the teacher model checkpoints are from the official publication.
% There is no doubt that a larger teacher model could bring more benefits.
When utilizing CLIP ViT-L/14 as a teacher, we set the input image resolution to 196$\times$196 for the teacher to match the number of patches with student ViT-B/16 or ViT-L/16.
As for the student model, we use the ViT-Base/Large equipped relative positional embeddings and layer scale mechanism following BEiT~\citep{beit, beitv2}.
For the pretraining setting, we mainly follow BEiT~\citep{beit, beitv2}: batch size 2048, learning rate 1.5e-3, AdamW optimizer with weight decay 0.05, drop path 0.1 (0.2) for ViT-Base(large), block-wise mask 40\%, epochs 300/800. More details can be found in Appendix.

\textbf{Evaluation.}
We consider the popular evaluating protocol for image classification on ImageNet-1k dataset: \textit{fine-tuning} top-1 accuracy. 
We adopt the BEiT~\citep{beit} fine-tuning recipe: 
For ViT-Base, we fine-tune it for 100 epochs with 20 epochs warm-up, and use AdamW optimizer with weight decay 0.05, learning rate 5e-4, and decays in a cosine schedule, layer decay 0.65;
For ViT-Large, we fine-tune it for 50 epochs with 5 epochs warm-up, layer decay 0.75.
For ViT-Huge, we fine-tune it for 30 epochs with 5 epochs warm-up, layer decay 0.85.
All the resolutions of input images are $224\times 224$.

As for the semantic segmentation task, we evaluate the \textit{mIoU} metric on ADE20K dataset~\citep{ade20k} with UperNet~\citep{upernet} framework. The input image resolution for training and evaluating are $512\times 512$. 
Remarkably, for the ViT-H/14 in Table~\ref{tbl:results:cls:imagenet_2}, we convert it to ViT-H/16 for sementic segmentation task.
Similarly, AdamW optimizer with weight decay of 0.05 is applied. Additionally, the training steps are 160K, and the batch size is 16.
And we employ learning rate \{5e-5, 8e-5, 1e-4\}, layer decay 0.75 (0.85), drop path 0.1 (0.2) for ViT-Base (Large).
More details can be found in Appendix.

\subsection{Main Results}

Table~\ref{tbl:results:cls:imagenet} reports the top-1 accuracy of some self-supervised methods on ImageNet-1k using ViT~\citep{vit} models. 
% We compare \our{} with recent popular supervised and self-supervised methods: 
% 1) supervised methods, like ViT~\citep{vit} and DeiT series ~\citep{deit,deit3}; 
% 2) contrastive-based methods, like MoCo v3~\citep{mocov3}, DINO~\citep{dino} and CLIP~\citep{clip};  
% 3) MIM-based methods, mentioned in Table~\ref{tbl:unify} 
% and 
% 4) hybrid methods containing contrastive learning and MIM, like  SIM~\citep{splitmask,sim}.
% With the same pretraining schedule, \our{} outperforms the latest supervised methods \deitiii{} by 1.7\% and 2.2\% upon ViT-Base and ViT-large respectively.
% \our{} also can consistently outperform contrastive-based methods, and boost them when taking them as teachers, refer to Table~\ref{tbl:results:abla:teacher}.
% Notably, following the BEiT~\citep{beit} fine-tuning recipe, CLIP~\citep{clip} can reach 84.9\% top-1 accuracy upon official ViT-Base checkpoint.
% Using it as teacher, \our{} can reach 85.5\% accuracy and 53.7\% mIoU upon ViT-Base, 87.1\% accuracy and 56.5\% mIoU upon ViT-large, on fine-tuning and segmentation task respectively.
% After scaling up to the larger teacher, like CLIP-L/14, \our{} can reach 88.3\% accuracy on ImageNet-1k using the ViT-H/14 backbone. 
% Compared with pixel-based and EMA-based target features, it substantially boosts the performance of masked image modeling.
For ViT-base, \our{} with 800 epochs pretraining schedule obtains 85.5\% top-1 accuracy, surpasses CLIP~\citep{clip}, MVP~\cite{mvp}, data2vec~\citep{data2vec} and MaskFeat~\citep{maskfeat} by 0.6\%, 1.1\%, 1.3\% and 1.5\% respectively. And \our{} also achieves comparable performance with \beitii{}~\citep{beitv2} on ImageNet-1k but outperforms \beitii{} by 1.2\% mIoU on ADE20k. More comparison with \beitii{} can be found in Section~\ref{sec:sec:compare_beit2}.
When scaling up the student to ViT-Large, \our{} achieves 86.8\% top-1 accuracy and 56.3\% mIoU. Compared to the recently proposed MILAN~\citep{MILAN}, \our{} outperforms it by 1\% on the semantic segmentation task under the less pretraining epochs.

In Table~\ref{tbl:results:cls:imagenet_2}, we use the CLIP ViT-Large/14 checkpoint as the teacher model and pretrain student models for 300 epochs. 
One can see that \our{} can get consistent improvements compared to teacher CLIP ViT-Base/16.
Remarkably, \our{} can reach 88.3\% accuracy on ImageNet-1k and 58.8\% mIoU on ADE20k by using the ViT-Huge backbone.

\begin{table*}[t]
\centering
\small
\begin{minipage}{3.1in}
\centering
\caption{Robustness evaluation on ImageNet variants~\citep{adversarial2021,rendition2021,sketch2019}.
}
\label{tbl:results:robust}
\resizebox{\linewidth}{!}{ % resize
\begin{tabular}{@{}lccc@{}}
\toprule
\bf \multirow{2}{*}{Methods} & \bf ImageNet & \bf ImageNet & \bf ImageNet \\
 & \bf Adversarial & \bf Rendition & \bf Sketch \\
\midrule
\multicolumn{4}{l}{\textit{ViT-B/16}} \\
MAE  & 35.9 & 48.3 & 34.5 \\
\beitii{}  & \textbf{54.4} & 61.0 & 45.6 \\
\our{}  & 53.3 & \textbf{64.4} & \textbf{47.3} \\
\midrule
\multicolumn{4}{l}{\textit{ViT-L/16}} \\
MAE &  57.1 & 59.9 & 45.3 \\
\beitii{} & \textbf{69.0} & 69.9 & 53.5 \\
\our{} & \textbf{69.0} & \bf 75.3 & \bf 56.9 \\
\bottomrule
\end{tabular}
} % resize
\end{minipage}
\hfill
\begin{minipage}{3.3in}
\centering
\caption{\our{} $vs$ knowledge distillation. The teacher model is CLIP ViT-Base~\citep{clip}. 
}
\label{tbl:results:compare_kd}
\resizebox{\linewidth}{!}{ % resize
\begin{tabular}{@{}c|cc|c@{}}
\toprule
\bf Student & \bf Mask & \bf Pretaining & \bf Classification  \\
\bf Models & \bf Ratios & \bf Epochs & \bf Accuracy (\%) \\
\midrule
\multirow{2}{*}{ViT-B/16} & 0 & 300  & \bf 85.3 \\
& 40\% & 300  & 85.0 ({\color{red} -0.3}) \\ 
\midrule
\multirow{2}{*}{ViT-B/16} & 0 & 800  & 85.2 \\
& 40\% & 800  &\bf 85.5 ({\color{blue} +0.3}) \\
\midrule
\multirow{2}{*}{ViT-L/16} & 0 & 300  & 85.4 \\
& 40\% & 300  &\bf 86.8 ({\color{blue} +1.4}) \\
\bottomrule
\end{tabular}
} % resize
\end{minipage}
% \vspace{-2em}
\end{table*}

\paragraph{Robustness evaluation.}
% We evaluate the robustness of \our{} on various ImageNet validation sets, $\ie{}$, ImageNet-Adversarial~\citep{adversarial2021}, ImageNet-Rendition~\citep{rendition2021} and ImageNet-Sketch~\citep{sketch2019}.
% As shown in Table~\ref{tbl:results:robust}, compared with MAE~\citep{mae}, \our{} achieves dramatic gains across datasets, demonstrating the superiority of the proposed method in terms of model generalization.
Following MAE~\citep{mae} and \beitii{}~\citep{beitv2}, we test the robustness of \our{} on three ImageNet validation sets, $\ie{}$, ImageNet-Adversarial~\citep{adversarial2021}, ImageNet-Rendition~\citep{rendition2021} and ImageNet-Sketch~\citep{sketch2019}.
In Table~\ref{tbl:results:robust}, both MAE and \beitii{} pretrain 1600 epochs, while \our{} pretrains 800 epochs but achieves comparable or superior performance. 

\subsection{Comparison with Knowledge Distillation}

% \begin{table}[t]
% \centering
% % \resizebox{\linewidth}{!}{ %resize
% \begin{tabular}{@{}c|cc|c@{}}
% \toprule
% \bf Student models $\gS$ & \bf Mask ratios & \bf Pretraining epochs & \bf Classification Accuracy(\%) \\
% \midrule
% \multirow{2}{*}{ViT-B/16} & 0 & 300  & 85.3 \\
% & 40\% & 300  & 85.0 ({\color{red} -0.3}) \\ 
% \midrule
% \multirow{2}{*}{ViT-B/16} & 0 & 800  & 85.2 \\
% & 40\% & 800  & 85.5 ({\color{green} +0.3}) \\
% \midrule
% \multirow{2}{*}{ViT-L/16} & 0 & 300  & 85.4 \\
% & 40\% & 300  & 86.8 ({\color{green} +1.4}) \\
% \bottomrule
% \end{tabular}
% % } % resize
% \caption{\our{} $vs$ knowledge distillation. The teacher model is CLIP ViT-Base here.
% }
% \label{tbl:results:compare_kd}
% \end{table}

In Table~\ref{tbl:results:compare_kd}, we compare \our{} with  knowledge distillation, which can be considered as a special case of \our{} where the mask ratio is 0 and loss is calculated on all patches. Knowledge distillation surpasses \our{} by 0.3\% when the pretraining schedule is 300 epochs, but is inferior to \our{} by 0.3\% when the pretraining schedule is 800 epochs.
Remarkably, \our{} outperforms knowledge distillation by a significant gain when the student model scales up to large-size models. 
The commonly used teacher model is CLIP ViT-Base, which reaches 84.9\% fine-tuning accuracy in terms of image classification on ImageNet-1k.
% That is, with the same model size as the teacher, both knowledge distillation and \our{} can outperform the teacher model itself, which is consistent with \citep{fd_clip}.

When the student is larger than the teacher, the student is easy to fully reconstruct the latent space of the teacher without information bottleneck.
This is why ViT-L/16 obtains comparable performance with ViT-B/16 (85.4\% $vs$ 85.3\% in Table~\ref{tbl:results:compare_kd}).
But in \our{}, under the condition of the corrupted input, the student is encouraged to extrapolate the masked patches, rather than mimicking features at visible patches.
% Therefore, the results of ViT-Large/16 in Table~\ref{tbl:results:compare_kd} indicate that predicting task itself drives the large student to learn, not just the mimicking.

% When the student is larger size, knowledge distillation struggles into fully mimic the output of teacher with full input image, leading to overfitting and poor performance.
% But in \our{}, under the condition of the corrupted input, model is encouraged to extrapolate the masked regions 
% it can alleviate this problem and learn a larger size model from a small teacher model.

\subsection{Comparison with \beitii{}}
\label{sec:sec:compare_beit2}

In \beitii~\citep{beitv2}, CLIP ViT-Base as the teacher model is responsible for distilling a vector quantized visual tokenizer, which provides the supervision for the subsequent MIM phase.
But compared with \our{}, the quantized mechanism in \beitii{} omits some fine-grained details from the teacher model.
And these details are beneficial to the fast convergence of \our{}, \eg{}, \our{} achieves comparable image classification performance with 800 epochs pretraining while \beitii{} need to pretrain 1600 epochs, as demonstrated in Table~\ref{tbl:results:cls:imagenet}.
That is, \our{} can avoid the codebook collapse problem in the tokenizer training~\citep{beitv2} but achieve comparable performance.
Meanwhile, such fine-grained details as supervision enhance the robustness of \our{}, as shown in Table~\ref{tbl:results:robust}.

% As demonstrated in \citep{fd_clip}

% TODO: re-organize
% In Table~\ref{tbl:results:compare_kd}, we also compare the masked feature distillation method with the recent knowledge distillation method~\citep{fd_clip}, when the teacher model is DINO and CLIP. In terms of DINO, the teacher model can achieve 83.6\% top-1 fine-tuning accuracy and 46.8\% mIoU for segmentation task. After feature distillation, FD-DINO~\citep{fd_clip} outperforms DINO by 0.2\% and 0.9\% on fine-tuning and segmentation respectively. However, after masked distillation, the student can surpass the teacher by 0.9\% and 3.6\% on fine-tuning and segmentation respectively.
% Moreover, when the teacher is CLIP~\citep{clip}, \our{} can consistently exceed the teacher model and show the scalability.
% However, after masked distillation, the student performs mostly different from the teacher: 
% need a metric
% 1) \textit{\color{red} Qualitatively}, from Figure~\ref{fig:maskdistill}(b), attention distribution between student and teacher is different. The reference patch in student pays more attention on others than itself, whether the input image is masked or not. Because the student model in pretraining process requires to capture more context information from others to predict itself.
% 2) \textit{Quantitatively}, from Table~\ref{tbl:results:cls:imagenet}, student from \our{} substantially boost the teacher model by 0.9\% on fine-tuning and 2.6\% on segmentation.

\subsection{Ablation Studies}
\label{sec:sec:abla}

\paragraph{Teacher models.}
We collect some popular unsupervised models to act as the teacher in \our{}, and pretrain a student model ViT-Base for 300 epochs in a MIM fashion. The performance of the teacher and student are shown in Table~\ref{tbl:results:abla:teacher}. 
From \#1 to \#6, where teacher models are CLIP and SLIP~\citep{slip} trained on the image-text pair datasets (YFCC15M, CC3M, CC12M and private 400M) in a language-guided contrastive way, \our{} consistently boost the teacher model by 0$\sim$3.3\% accuracy.
From \#7 to \#8, teacher models SimCLR~\citep{simclr} and DINO~\citep{dino} only use image data. \our{} boosts them by 1.6\% and 0.9\% respectively.

% \textit{Does language information really bring substantial gains for MIM?} 
Comparing \#1, \#2 and \#7 in Table~\ref{tbl:results:abla:teacher}, where the same dataset and training epochs are applied to teachers, students in \#1 and \#7 respectively achieve 83.8\% and 84.1\%, but the former using the text information and the later not, implying that the language-guided supervision is not essential.
Moreover, comparing \#1$\sim$\#5 and \#8, both teacher and student in \#8 trained on ImageNet-1k can reach comparable performance with those in \#1$\sim$\#5, which further suggests that the extra language information is not the key.
% some conclusion
% balabala

% \textit{Are only contrastive pretrained models suitable to be teacher for MIM?} 
From \#9 to \#11, we choose the model trained by MIM itself to act as the teacher model. We find that \our{} consistently outperform the corresponding teacher.
However, comparing \#8 with \#9, where teacher can reach the same fine-tuning accuracy, students in \#8 can obtain better performance in terms of fine-tuning accuracy and segmentation mIoU than those in \#9, indicating that contrastive pretrained models tend to be the better but not the only solution.

\begin{table}[!t]
\centering
\caption{Ablation studies on teacher models used in \our{}. For \{CLIP, SLIP, SimCLR\}$^{\ddagger}$, the fine-tuning accuracy and model checkpoint are all from SLIP~\citep{slip}. For CLIP$^{*}$ and DINO, we use the official model checkpoint and follow BEiT~\citep{beit} fine-tuning recipe to get the top-1 accuracy. The teacher models in all methods are \textit{ViT-Base} model. The student model is ViT-Base and is pretrained for 300 epochs.
}
\label{tbl:results:abla:teacher}
\begin{tabular}{@{}l|lccc|cc@{}}
\toprule
& \multicolumn{4}{c|}{\textbf{Teacher Model} $\gT$} &  \multicolumn{2}{c}{\textbf{Student Model} $\gS$} \\ 
& \bf Teacher & \bf Data & \bf Text & \bf ImageNet (\%) &\bf ImageNet (\%) &\bf ADE20k (\%) \\
\midrule
\#1 & CLIP$^{\ddagger}$  & YFCC15M & \cmark  &  80.5 & 83.8 ({\color{blue} +3.3})& 47.4 \\
\#2 & SLIP$^{\ddagger}$  & YFCC15M & \cmark  &  82.6 & 84.3 ({\color{blue} +1.7}) & 49.9 \\
\#3 & SLIP$^{\ddagger}$  & YFCC15M & \cmark  &  83.4 & 84.6 ({\color{blue} +1.2}) & 50.8 \\
\#4 & CLIP$^{\ddagger}$  & CC3M & \cmark  & 79.5 & 83.7 ({\color{blue} +4.2}) & 45.7 \\
\#5 & CLIP$^{\ddagger}$  & CC12M & \cmark  & 82.1 & 84.1 ({\color{blue} +2.0}) & 48.3 \\
\#6 & CLIP$^{*}$  & Private 400M & \cmark  &  84.9 & \textbf{85.0} ({\color{blue} +0.1}) & \textbf{53.8} \\
\midrule
\#7 & SimCLR$^{\ddagger}$  & YFCC15M & {\color{magenta} \xmark}  &  82.5 & 84.1 ({\color{blue} +1.6}) & 49.4 \\
\#8 & DINO  & ImageNet-1k & {\color{magenta} \xmark}  &  83.6 & 84.5 ({\color{blue} +0.9}) & 50.4 \\
\midrule
\#9 & MAE  & ImageNet-1k & {\color{magenta} \xmark}  &  83.6 & 84.3 ({\color{blue} +0.7}) & 49.3 \\
\#10 & BEiT  & ImageNet-1k & {\color{magenta} \xmark}  &  83.2 & 83.8 ({\color{blue} +0.6}) & 46.6 \\
\#11 & BEiT v2  & ImageNet-1k & {\color{magenta} \xmark}  &  84.7 & \textbf{85.0} ({\color{blue} +0.3}) & 52.1 \\
\bottomrule
\end{tabular}
\end{table}

\paragraph{Loss functions \& Normalization.}
We compare MSE, cosine similarity and smooth-$\ell_1$ loss equipped with various normalization layers, then present the results in Table~\ref{tbl:results:abla:loss}. 
% MSE loss means that the outputs of student require to fully mimic those of teacher, but cosine similarity and smooth-$\ell_1$ loss just needs student to mimic the teacher outputs after $l2$ and $\ell_1$ normalization respectively, which relaxes the student in some extend.
From Table~\ref{tbl:results:abla:loss}, one can see that smooth-$\ell_1$ loss equipped with LN can achieve better performance under the supervision of both DINO and CLIP, indicating that Normalization plays an important role in masked image modeling task.

\paragraph{Target layer selection.}
Usually, the deeper layer feature of a model is biased to the special task, $e.g.$, image-image contrastive learning for DINO and image-text contrastive learning for CLIP. 
But whether it is beneficial for \our{} is not revealed. 
We conduct experiments on target feature from last layer, average of last 3 layers and average of last 6 layers.
As shown in Table~\ref{tbl:results:abla:feature_selection}, the last layer's features are better for DINO teachers while the last 6 layers' features are better for CLIP teachers.
% , implying that the better selection depends on the type of teacher.
Moreover, results on the segmentation task show that the last layer features as target are superior.
Therefore, we choose the last layer feature as the default target feature for all experiments.

\begin{figure}[t]
\begin{center}
\begin{tabular}{c}
\includegraphics[width=1\textwidth]{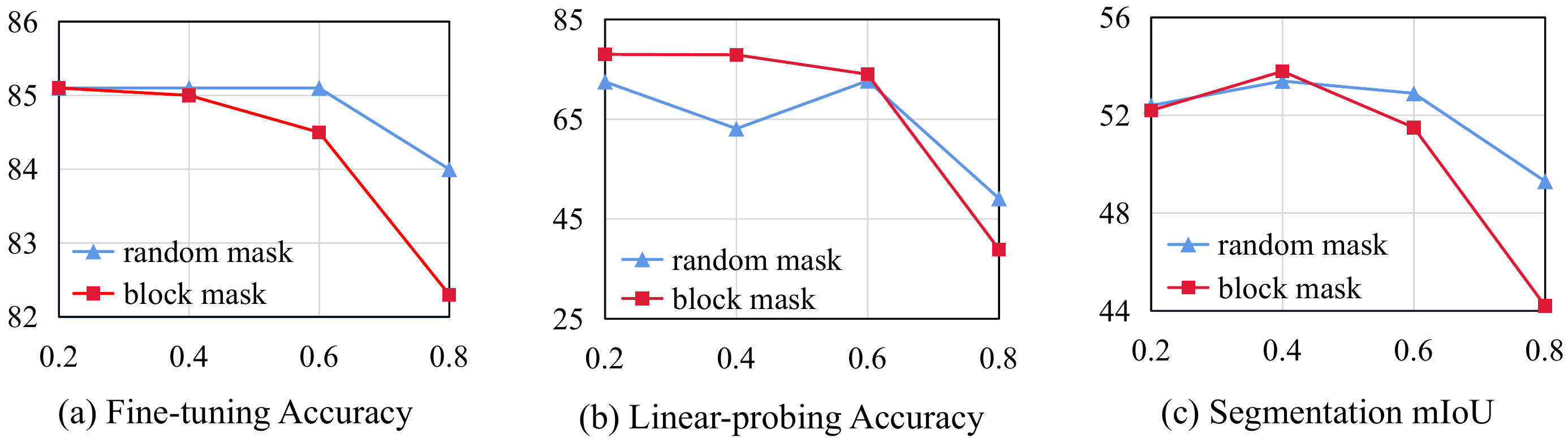}
\end{tabular}
\end{center}
\caption{The block-wise mask $vs$ random mask, under various mask ratios.
}
\label{fig:abla:mask_way}
\end{figure}

\paragraph{Masked strategy.}
For the masked strategy, we evaluate the block-wise~\citep{beit} masked method and random masked method in Figure~\ref{fig:abla:mask_way}. The block-wise masked method performs better than random mask under low mask ratios, while worse than random mask under high mask ratios.
Taking the three evaluation protocols (fine-tuning on ImageNet-1k, linear-probing ImageNet-1k, and semantic segmentation on ADE20k) into consideration, we choose the block-wise mask with 40\% mask ratio as the final decision.

% 

% \paragraph{Teacher input.}

% \paragraph{Iterative boost.}
% In Table~\ref{tbl:results:abla:teacher}, the student can outperform the corresponding teacher from various pretext tasks.
% Then a natural question arise: \textit{How about using the student as the teacher iteratively?}
% We investigate it in the Table~\ref{tbl:results:abla:iterative}.
% For \#1, it is the basic setting of BEiT v2 without cls-token pretraining, under the supervison of VQ-KD. The supervision of VQ-KD is also from CLIP here.
% Then, we use the student model in \#1 as the teacher model for \#2. After that, the student model can outperform the teacher by 0.3\%.
% Similarly, we use the student model in \#4 as the teacher model for \#3, but find it can't boost.
% Revisiting \#1 and \#2, the supervision in the iterative process is distinct  while that in \#3 and \#4 is consistent.
% Therefore, the target feature in \#5 is from the last 6 layers, result in 0.3\% gain.
% However, how to iterative boost itself is not the purpose of this work, and we leave it in the future.

\begin{table*}[t]
\centering
\small
\begin{minipage}{3.2in}
\centering
\caption{Ablation study of loss functions and normalization layers. All models are pretrained for 300 epochs. 
}
\label{tbl:results:abla:loss}
\begin{tabular}{@{}lcccc@{}}
\toprule
\bf $\gT$  & \bf $\gL$ & \bf Norm & \bf ImageNet & \bf ADE20k \\
\midrule
\multirow{3}{*}{DINO}  & MSE & \xmark & 84.3 & 49.6 \\
  & Cosine & ($\ell_2$) & 84.5 & 49.6 \\
  & Smooth-$\ell_1$ & LN & 84.5 & 50.4 \\
\midrule
\multirow{3}{*}{CLIP}  & MSE & \xmark & 84.6 & 52.8 \\
 & Cosine & ($\ell_2$) & 84.9 & 52.9 \\
 & Smooth-$\ell_1$ & BN & 84.9 & 53.1 \\
 & Smooth-$\ell_1$ & LN & 85.0 & 53.8 \\
\bottomrule
\end{tabular}
\end{minipage}
\hfill
\begin{minipage}{3.2in}
\centering
\caption{Ablation study of target feature selection in \our{}. All models are pretrained for 300 epochs. 
}
\label{tbl:results:abla:feature_selection}
\begin{tabular}{@{}llcc@{}}
\toprule
\bf $\gT$  & \bf Target & \bf ImageNet & \bf ADE20k \\
\midrule
\multirow{3}{*}{DINO}  & Last layer & 84.5 & 50.4 \\
  & Mean (last 3 layers)  & 84.4 & 49.7 \\
  & Mean (last 6 layers)  & 84.3 & 49.8 \\
\midrule
\multirow{3}{*}{CLIP}  & Last layer & 85.0 & 53.8 \\
  & Mean (last 3 layers)  & 85.0 & 53.5 \\
  & Mean (last 6 layers)  & 85.1 & 53.4  \\
\bottomrule
\end{tabular}
\end{minipage}
% \hfill
% \begin{minipage}{0.5\linewidth}
% \centering
% \begin{tabular}{@{}l|cccc@{}}
% \toprule
% & \bf $\gT$ & \bf Loss & \bf Target & \bf FT \\
% \midrule
% \#1 & VQ-KD & CrossEntropy & Visual code & 84.7 \\
% \#2 & $\gS$ in \#1 & Smooth-$\ell_1$ & Last layer & 85.0 \\
% \midrule
% \#3 & CLIP & Smooth-$\ell_1$ & Last layer  & 84.9 \\
% \#4 & $\gS$ in \#3 & Smooth-$\ell_1$ & Last layer & 84.9 \\
% \#5 & $\gS$ in \#3 & Smooth-$\ell_1$ & Last 6 layers  & 85.2 \\
% \bottomrule
% \end{tabular}
% \caption{Iterative boost. $\gS$ in \#\{2, 4, 5\} initializes from scratch, not the teacher, in the iterative process. 300 epochs schedule is applied.} 
% \label{tbl:results:abla:iterative}
% \end{minipage}
\end{table*}

% \section{What properties does \our{} change?}
\subsection{Analysis: MIM Enhances Shape Bias}

\begin{figure}[t]
\centering
\subfloat{
\includegraphics[width=0.46\textwidth]{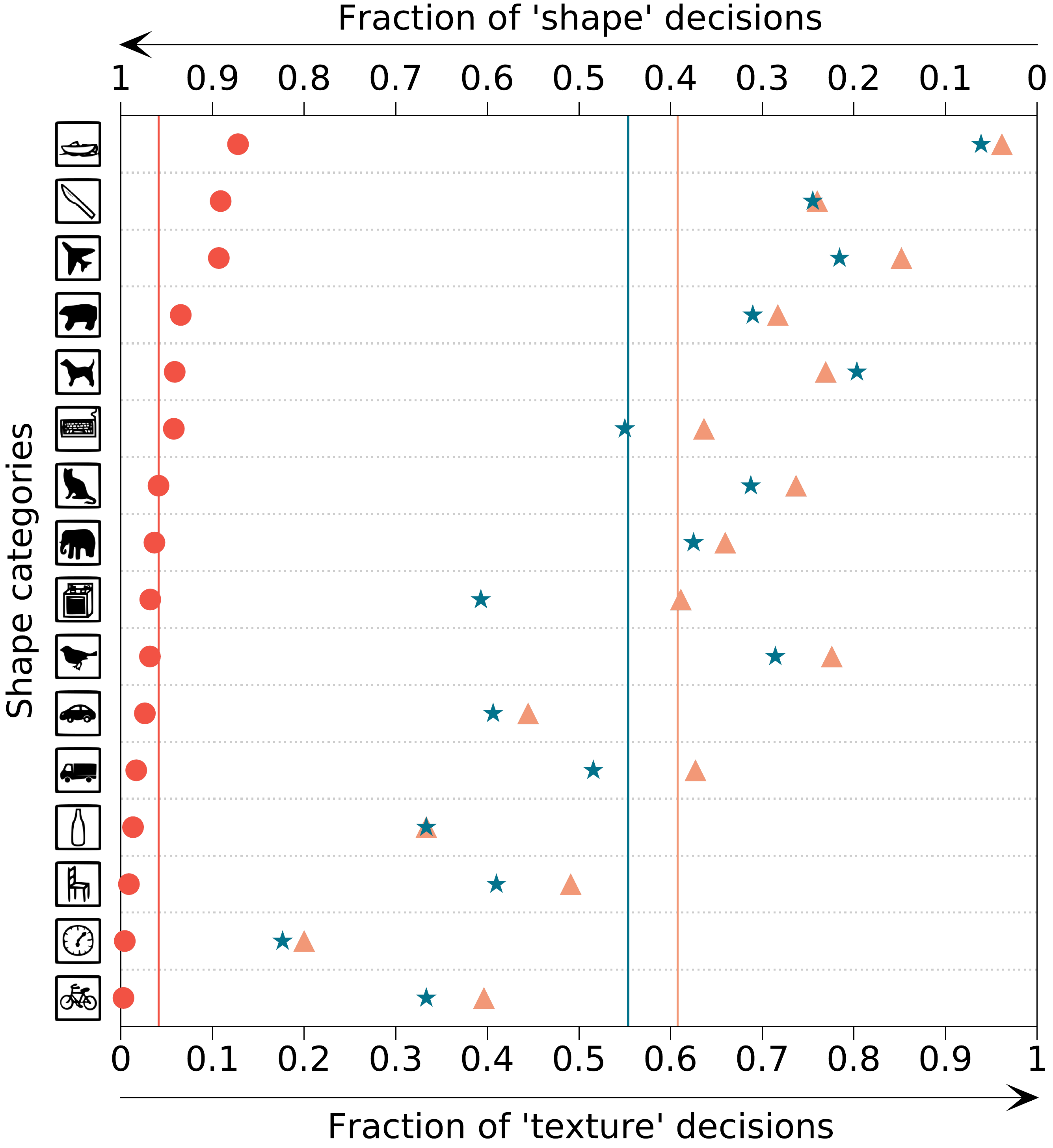}
}
% \hfill
\hspace{0.1in}
\subfloat{
\includegraphics[width=0.46\textwidth]{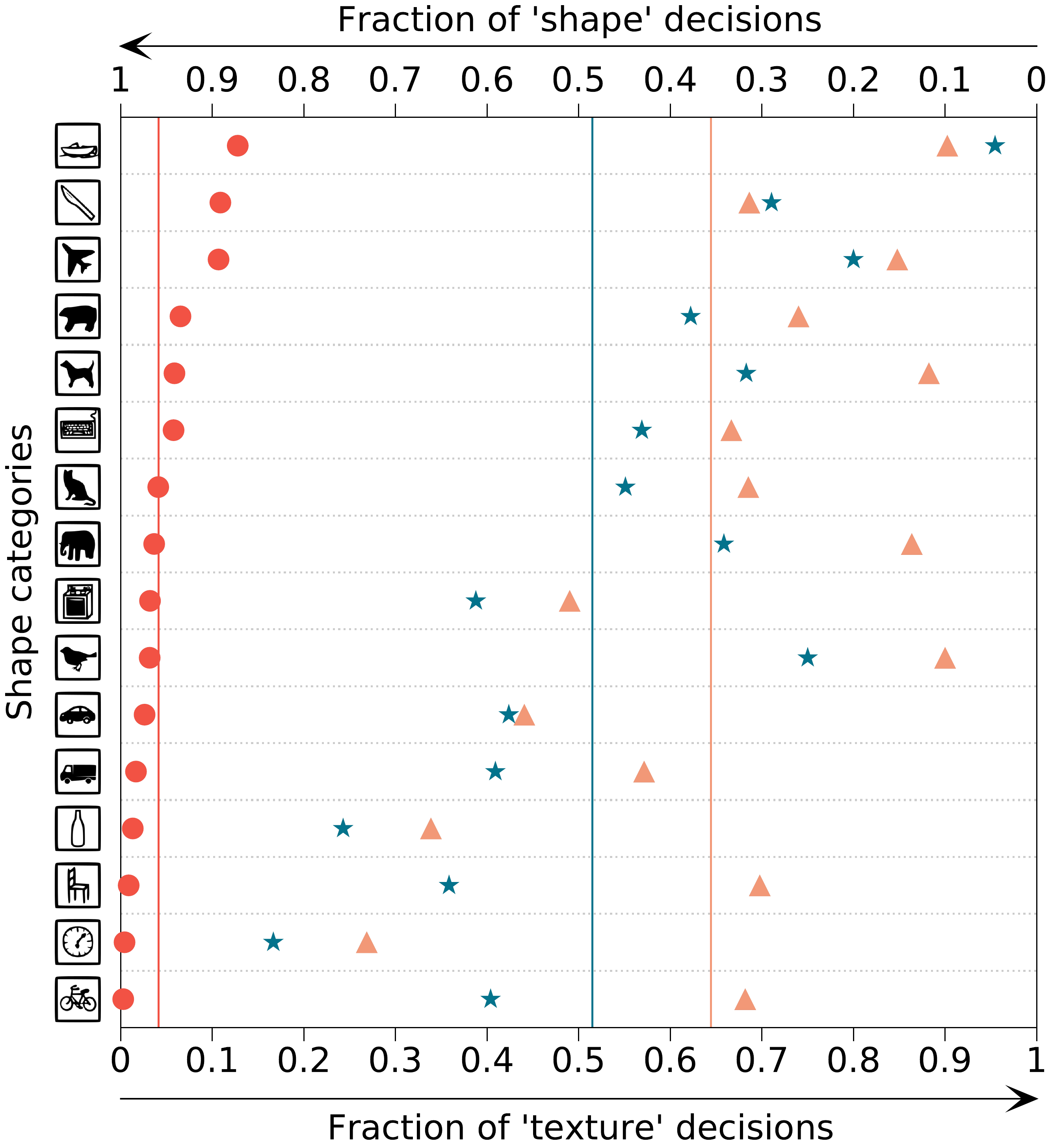}
}
\caption{Shape-biased analysis under the teacher supervision of CLIP ViT-B/16 (\textbf{left}) and MAE ViT-B/16 (\textbf{right}). {\color{human} Circle}, {\color{teacher} triangle} and  {\color{student} star} denote humans, teachers and students, respectively. Vertical lines are the corresponding average values. Masked image modeling enhances the shape bias. (Best viewed in color)
}
\label{fig:shape_bias}
\end{figure}

% In addition to better performance, what influence MIM has brought is a question that needs to answer.
% Compared with convolutional neural networks (CNNs), the robustness of ViTs is not due to the texture bias, but the shape bias~\citep{naseer2021intriguing}.
% After being properly trained to encode shape features, ViTs even can obtain the same shape recognition ability as the human visual system.
% And ViT models outperform convolutional neural networks in terms of shape bias~\citep{naseer2021intriguing}.
We explore whether the masked image modeling methods can enhance the shape-biased ability or not.
The fraction of correct decisions based on object shape is characterized as shape bias.
\citet{naseer2021intriguing} present that human usually is much more shape-biased compared with supervised classification models, such as convolutional networks, and vision Transformers.
We evaluate the shape bias capacity on a stylized version of ImageNet~\citep{naseer2021intriguing} by using the checkpoints fine-tuned on the original ImageNet-1k dataset.
As shown in Figure~\ref{fig:shape_bias}, masked image modeling tends to promote the shape bias of the models.
The results partially explains why \our{} generalizes better on ImageNet variants as shown in Table~\ref{tbl:results:robust}.
% Interestingly, the student under the supervision of MAE outperforms the student under the supervision of CLIP, indicating that the teacher trained with the masked image modeling method is more shape-biased than contrastive-based methods, and \our{} can further enhance the shape-biased ability for student models.

\section{Related Work}

\paragraph{Masked image modeling.}
Masked language modeling task root in Transformers has achieved great success in learning strong language representations in recent years~\citep{bert,unilm,unilm2}.
Inspired by it, \beit{}~\citep{beit} proposes a mask-then-predict framework to recover discrete visual tokens~\citep{dalle}, which shows the great potential of masked image modeling for the computer vision field.
% After that, the components of masked image modeling are explored by many recent works.
% MAE~\citep{mae} considers MIM as a denoising pixel-level reconstruction task.
% \beitii{}~\citep{beitv2} introduces \vqkd{} to construct a semantic-rich target for \beit{}. 
After that, various target supervision has been explored under the masked image modeling framework, such as original or normalized pixels~\citep{mae,peco,swinv2,ConvMAE,Ge2-AE,HiViT,GreenMIM}, high-level features~\citep{maskfeat,mvp,beitv2,MimCo,MILAN}, and EMA-updated models~\citep{data2vec,MSN,sim,SdAE,ConMIM,ExtreMA,BootMAE}.
% MAE~\citep{mae}, SimMIM~\citep{swinv2}, ConvMAE~\citep{ConvMAE}, Ge$^2$-AE~\citep{Ge2-AE}, HiViT~\citep{HiViT} and GreenMIM~\citep{GreenMIM} consider original or normalized pixel as reconstruction targets while SimMIM~\citep{swinv2}, ConvMAE~\citep{ConvMAE}, HiViT~\citep{HiViT} and GreenMIM~\citep{GreenMIM} apply MIM on ViT variants.
% Meanwhile, taking the semantic richness of MIM targets into account, PeCo~\citep{peco} uses MoCo v3~\citep{mocov3} as a perceptual model when training the tokenizer while \beitii{}~\citep{beitv2} constructs a tokenizer by distilling the CLIP model.
% And MaskFeat~\citep{maskfeat}, MVP~\cite{mvp}, MILAN~\citep{MILAN} and MimCo~\citep{MimCo} regress the feature of supervised or self-supervised models.
% EMA-updated models is also a alternative choice and extensively studied in data2vec~\citep{data2vec}, MSN~\citep{MSN}, SIM~\citep{sim}, sdAE~\citep{SdAE}, ConMIM~\citep{ConMIM}, ExtreMA~\citep{ExtreMA} and BootMAE~\citep{BootMAE}.
% highlight that MILAN is a concurrent work with ours.
% As a concurrent work, MILAN~\citep{MILAN} also proposes to regress CLIP features with a special decoder and mask strategy. 
In this work, we decouple and analyze the components of the recent masked image modeling works, and then propose a simple yet effective paradigm for masked image modeling.

\paragraph{Contrastive learning.}
As a simple but effective self-supervised method, contrastive learning methods have ushered in rapid progress in recent years.
The main idea is to enforce similarity over augmented views of an image and push the views augmented from other images away~\citep{dosovitskiy2016discriminative,wu2018unsupervised,dim,moco,simclr}, or to avoid model collapse after removing negative pairs~\citep{byol,simsiam,mocov3,dino}.
In the multimodal field, CLIP~\citep{clip} and ALIGN~\citep{align} can learn image-language alignment representation, by grouping positive image-text pairs (an image and corresponding tag or caption) closer and separating negative image-text pairs. 
And SLIP~\citep{slip} combines language supervision and image self-supervision to further boost the learned visual representations.
In this work, we consider contrastive models as the target for masked image modeling.

\paragraph{Knowledge distillation.}
Knowledge distillation~\citep{hinton2015distilling} considers the output of the teacher model as the pseudo label to learn the student model. Such a strategy squeezes the potential of small models and brings impressive gains. After that, knowledge distillation is transferred to various tasks~\citep{deit,He2019KnowledgeAF,Yang2021KnowledgeDV} and domains~\citep{TinyBert,MiniLM}.
{\cite{fd_clip} proposes that using the normalized feature from teacher fully distills a same size student.}
However, in this work, \our{} aims to reconstruct the corresponding teacher output at masked patches rather than mimicking the teacher's feature at each patch.

\section{Conclusion and Limitations}

We summarized the existing MIM works upon the proposed unified view: teacher models, student models, normalization layers and MIM heads.
After that, we propose a simple yet effective method, termed as \our{}, which predicts the normalized semantic features from CLIP's visual encoder at masked positions based on the corrupted input image.
The simple framework beats many previous works with special designs and shows impressive performance across model sizes and tasks.
In the future, we would like to explore the proposed method for multimodal pretraining~\citep{beit3}.
% the apposite supervision for MIM task, \ie{}, what supervision MIM task demands.

% \section*{Limitations}
The proposed \our{} requires an extra teacher model, similar to the tokenizer in \beit{} series.
Compared with the methods using pixels as targets, the teacher model in \our{} needs to spend extra time to obtain target features.
Meanwhile, we point out that language-guided supervision is not essential in Susection~\ref{sec:sec:abla} on the academically accessible multi-model datasets, YFCC15M.
But whether this conclusion is correct on private 400M image-text pair datasets remains an unknown question.

\bibliographystyle{tmlr}
\bibliography{main}

\newpage

\appendix

\section{Hyperparameters for \our{} Pretraining}
\label{app:pretrain}

\begin{table}[h!]
\centering
\small
\scalebox{0.98}{
\begin{tabular}{l|ccc}
\toprule
\bf Hyperparameters & \bf Base Size & \bf Large Size & \bf Huge Size \\
\midrule
Layers & 12 & 24 & 32 \\
Hidden size & 768 & 1024 & 1280 \\
FFN inner hidden size & 3072 & 4096 & 5120 \\
Attention heads & 12 & 16 & 16 \\
Layer scale & 0.1 & 1e-5 & 1e-5 \\
Patch size & $16 \times 16$ & $16 \times 16$ & $14 \times 14$ \\
% Patch size & \multicolumn{2}{c}{$16 \times 16$} \\
% Relative positional embeddings & \multicolumn{2}{c}{\cmark} \\
% Shared relative positional embeddings & \multicolumn{2}{c}{\cmark} \\
\midrule
Training epochs & \multicolumn{3}{c}{300/800} \\
Batch size & \multicolumn{3}{c}{2048} \\
Adam $\epsilon$ & \multicolumn{3}{c}{1e-8} \\
Adam $\beta$ & \multicolumn{3}{c}{(0.9, 0.999)} \\
Peak learning rate & \multicolumn{3}{c}{1.5e-3} \\
Minimal learning rate & \multicolumn{3}{c}{1e-5} \\
Learning rate schedule & \multicolumn{3}{c}{Cosine} \\
Warmup epochs & \multicolumn{3}{c}{10} \\
\midrule
Stoch. depth & 0.1 & 0.2 & 0.25 \\
Gradient clipping & \multicolumn{3}{c}{3.0} \\
Dropout & \multicolumn{3}{c}{\xmark} \\
% Stoch. depth & \multicolumn{2}{c}{\xmark} \\
Weight decay & \multicolumn{3}{c}{0.05} \\
\midrule
Data Augment & \multicolumn{3}{c}{RandomResizeAndCrop} \\
Input resolution & \multicolumn{3}{c}{$224 \times 224$} \\
Color jitter & \multicolumn{3}{c}{0.4} \\
\bottomrule
\end{tabular}
}

\caption{
Hyperparameters for \our{} pretraining on ImageNet-1K.
}
\label{tbl:pretrain:hyperparams}
\end{table}

\section{Hyperparameters for ADE20K Semantic Segmentation Fine-tuning}
\label{app:finetune:seg}

\begin{table}[h!]
\centering
\begin{tabular}{l|c c}
\toprule
\bf Hyperparameters & \bf ViT-B/16 & \bf ViT-L/16 \\
\midrule
Relative positional embeddings & \multicolumn{2}{c}{\cmark} \\
Shared relative positional embeddings & \multicolumn{2}{c}{\xmark} \\
\midrule
Peak learning rate & \multicolumn{2}{c}{\{0.5, 0.8, 1.0, 1.5\}e-4} \\
% Peak learning rate & & \\
Fine-tuning steps & \multicolumn{2}{c}{160K} \\
Batch size & \multicolumn{2}{c}{16} \\
Adam $\epsilon$ & \multicolumn{2}{c}{1e-8}  \\
Adam $\beta$ & \multicolumn{2}{c}{(0.9, 0.999)} \\
Layer-wise learning rate decay & 0.75 & 0.85 \\
Minimal learning rate & \multicolumn{2}{c}{0} \\
Learning rate schedule & \multicolumn{2}{c}{Linear} \\
Warmup steps & \multicolumn{2}{c}{1500} \\
\midrule
Dropout & \multicolumn{2}{c}{\xmark} \\
Stoch. depth & 0.1 & 0.2 \\
Weight decay & \multicolumn{2}{c}{0.05} \\
\midrule
Input resolution & \multicolumn{2}{c}{$512 \times 512$} \\
\bottomrule
\end{tabular}
\caption{
Hyperparameters for fine-tuning \our{} on ADE20K.
}
\label{tbl:ft:ade20k:hyperparams}
\end{table}

\newpage
\section{Hyperparameters for Image Classification Fine-tuning}
\label{app:finetune:cls}

\begin{table}[ht]
\centering
\caption{
Hyperparameters for fine-tuning \our{} on ImageNet-1K.
}
\label{tbl:ft:imagenet:hyperparams}
\scalebox{0.95}{
\begin{tabular}{l|ccc}
\toprule
\bf Hyperparameters & \bf ViT-B/16 & \bf ViT-L/16 & \bf ViT-H/14 \\
\midrule
Peak learning rate & 5e-4 & 5e-4 & 2e-4\\
Fine-tuning epochs & 100  & 50 & 30\\
Warmup epochs & 20 & 5 & 5 \\
Layer-wise learning rate decay & 0.65 & 0.8 & 0.85\\
Batch size & \multicolumn{3}{c}{1024} \\
Adam $\epsilon$ & \multicolumn{3}{c}{1e-8}  \\
Adam $\beta$ & \multicolumn{3}{c}{(0.9, 0.999)} \\
Minimal learning rate & \multicolumn{3}{c}{1e-6} \\
Learning rate schedule & \multicolumn{3}{c}{Cosine} \\
\midrule
Stoch. depth & 0.1 & 0.2 & 0.25 \\
Repeated Aug & \multicolumn{3}{c}{\xmark} \\
Weight decay & \multicolumn{3}{c}{0.05} \\
Label smoothing $\varepsilon$ & \multicolumn{3}{c}{0.1}     \\
Dropout & \multicolumn{3}{c}{\xmark} \\
Gradient clipping & \multicolumn{3}{c}{\xmark} \\
\midrule
Erasing prob.  & \multicolumn{3}{c}{0.25} \\
Input resolution & \multicolumn{3}{c}{$224 \times 224$} \\
Rand Augment  & \multicolumn{3}{c}{9/0.5} \\
Mixup prob.  & \multicolumn{3}{c}{0.8}     \\
Cutmix prob.   & \multicolumn{3}{c}{1.0}    \\
\bottomrule
\end{tabular}
}
\end{table}

\end{document}